# ClinOCR-Bench: A Comprehensive Clinical Scanned Document Dataset for Optical Character Recognition Model Evaluation

**Enshuo Hsu[1, 2], MS; Jin Zhou[1], BS; Kirk Roberts[1], PhD**

[1]McWilliams School of Biomedical Informatics, University of Texas Health Science Center at Houston, Houston, Texas, USA.

[2]Enterprise Development & Integration, University of Texas MD Anderson Cancer Center, Houston, TX, USA.

**Corresponding author: Kirk Roberts, kirk.roberts@uth.tmc.edu**

**Word Count:**

<u>Total: 3146</u>

<u>Abstract: 170 (Max 170)</u>

## Abstract

Extracting textual information from scanned medical documents, such as external laboratory reports and manually filled forms, has been a major challenge in modern electronic health records (EHRs). Recent advancements in vision language models (VLMs) have shown great promise over traditional OCR tools. However, at this point, most clinical OCR studies were conducted on private, institutional data. To our knowledge, there are few publicly available datasets for evaluating OCR models in the clinical domain. Furthermore, common scanning artifacts that undermine OCR performance are not reflected in those datasets, leaving a systematic evaluation unfeasible. Therefore, we release a publicly available, realistic-looking OCR benchmark dataset, ClinOCR-Bench, with 384 scanned images across 6 subsets: *Normal*, *Handwriting*, *Poor Quality*, *Rotation*, *Tables*, and *Mix-artifacts*. ClinOCR-Bench features: 1) diverse document types and layouts, 2) full coverage of common EHR scan artifacts, 3) protected health information-free, 4) template-aware train/test split, and 5) adequate sample size for OCR benchmarking. Baseline OCR performance was evaluated using state-of-the-art open-weight and proprietary VLMs. The dataset and documentation are available on GitHub (https://github.com/ClinOCR-Bench/ClinOCR-Bench).

## Background & Summary

Scanned documents have been a long-standing challenge in electronic health record (EHR) systems, as the embedded information is unassessable for computational processing [1]. In 2018 alone, a single urban healthcare system received over 3.5 million scanned documents across 129 distinct document types [2]; while from 2015 to 2018, a single clinical department received over 3000 scanned reports for a single type of laboratory test [3]. The high-volume, diverse scanned documents have been a universal source of stress for EHR systems nationwide, causing challenges in efficiency and patient safety [4]. In current clinical workflows, scanned documents are often introduced from 1) previous medical records before transfer of care, 2) external laboratory reports, 3) manually filled forms, and 4) miscellaneous documents (e.g., insurance, faxes, and letters). To systematically process them, computer vision-based optical character recognition (OCR) engines (e.g., Tesseract) have been adopted [3,5]. Recent advancements in Vision Language Models (VLMs) have shown further improvement on general OCR benchmarks [6,7]. However, the performance is still far from perfect. Continuous model development and evaluation for the medical field are urgently needed.

Medical OCR is particularly challenging due to four common artifacts: 1) **Handwriting**, such as manually filled medical forms and laboratory reports [3,8]. OCR engines often fail to recognize handwritten letters, resulting in garbage characters (i.e., nonsensical and fragmented letters). 2) **Tabular data**, such as external lab or pathology reports [2]. OCR engines often capture content in rows but fail to align columns, resulting in a broken table layout in the output. This is especially problematic on nested tables. 3) **Poor-quality scanning**, including blurriness, fading text, and background noise [9]. Scanning with an older scanner, multi-generation copies (i.e., copying on copies), and scanning through sheet protectors could contribute to this artifact. 4) **Rotated or skewed scans**, often by 90, 180, or 270 degrees. OCR engines assume text is aligned horizontally. Rotation causes systematic misrecognition, leading to garbage characters.

Evaluating OCR models often requires large amounts of scanned documents with ground truth [10,11]. At this point, most medical OCR studies develop and evaluate models using institutional data, which is not publicly accessible. To our knowledge, there are few publicly available datasets for medical OCR. The *medical prescription dataset* [12] includes 1000 algorithm-generated handwritten prescriptions split into a training set (N=800), a validation set (N=100), and a test set (N=100). One layout template and one handwriting font are applied to all images. There is no blurriness, background noise, or rotation. The *Medical Lab Report Dataset* [13] comprises 426 images of de-identified laboratory reports. The main body of the reports consists of large tables with numeric values (e.g., complete blood count). The scanning quality is high-resolution and clear, with minimal handwriting. There is no ground truth associated with this dataset. The *OCR for Medical Laboratory Reports* [14] dataset consists of 238 images of scanned pathology reports, split into a training set (N=139), a validation set (N=47), and a test set (N=52). The reports are written in Chinese. All reports follow the same template and were scanned at a high resolution. There is no handwriting or skewness. The *African Medical Records (AMR)* [15] is a collection of manually created, synthetic handwritten medical records from contributors across 40 universities in Nigeria. At this point, it includes a total of 32 images across four types: patient visit notes, vital signs charts, prescription notes, and lab request forms. All images are well-photographed with cameras, fully visible, and written in clear and readable text. The *MedOCR Vision Dataset* [16] is a secondary dataset that adopts the *medical prescription dataset* [12] and the *Medical Lab Report Dataset* [13], along with two non-medical OCR datasets. The author generated a silver standard for the *Medical Lab Report Dataset* with Nebius and Novita image-to-text conversion APIs. In conclusion, current public medical OCR datasets generally lack diversity in layout (i.e., generated from a single template) and document types (i.e., mostly laboratory reports). Furthermore, the poor-quality scanning and rotation artifacts—which are common in EHR scans—were completely omitted.

Therefore, we propose ClinOCR-Bench, an original, realistic-looking dataset featuring: 1) **Diverse document types and layouts**. 16 manually created templates ranging from 12 document types (e.g., imaging report, patient intake form, clinical visit note, medication list, and referral letter) adopted from real medical documents, and 61 handwriting fonts, including a mixture of paragraphs, tables, and images, were applied to mimic the complexity of scanned EHR documents. 2) **Full coverage of common scan artifacts in EHR**. Our dataset comprises six subsets to support systematic OCR model development: normal quality (N=64), handwriting (N=64), tabular data (N=64), poor quality scanning (N=64), rotated scans (N=64), and mixed artifacts (N=64). 3) **Protected health information (PHI)-free**. ClinOCR-Bench is a completely synthetic dataset that can be shared anywhere without patient privacy restrictions. 4) **Template-aware train/test split**. Each image in the test set has a one-shot exemplar from the same template ("homogenous one-shot") and an exemplar from a different template ("heterogeneous one-shot") to enable consistent benchmarking under these typical learning scenarios. 5) **Adequate sample size**. Our test set includes 328 images across 6 subsets. With *Tesseract,* this yields a narrow 95% confidence interval for the overall word error rate (WER) with a margin of error below 4%. Our sample size is thus sufficient for a precise estimation of OCR performance. ClinOCR-Bench addresses current research and application gaps and enables the future development and evaluation of clinical OCR models (Figure 1).

Synthetic data has two major benefits over medical records: 1) It is completely PHI-free. As cloud computing (e.g., LLM inferencing through API calls) becomes increasingly adopted, many healthcare organizations and research institutes are outsourcing LLM inferencing to external vendors (e.g., OpenAI, Microsoft Azure, and OpenRouter). Synthetic data enables benchmarking of OCR systems anywhere, without restrictions. 2) It can achieve large sample size with accurate ground truth. Manual annotation of ground truth from clinical scanned documents is labor-intensive and error-prone. The synthetic workflow, in which textual content is available in digital files, makes ground truth creation far more efficient and thus allows for greater scale. Despite the benefits, a valid concern about synthetic data is its capacity to proxy for real-world medical records. Unfortunately, to our knowledge, there is no established best practice for scanned document synthesis. As a pioneering effort, we a) adopted layouts (i.e., templates) inspired by scanned medical documents; b) synthesized images, textual content, and tabular content with LLMs to create contextually sound documents; and c) adopted common artifacts to replicate the challenges in clinical OCR.

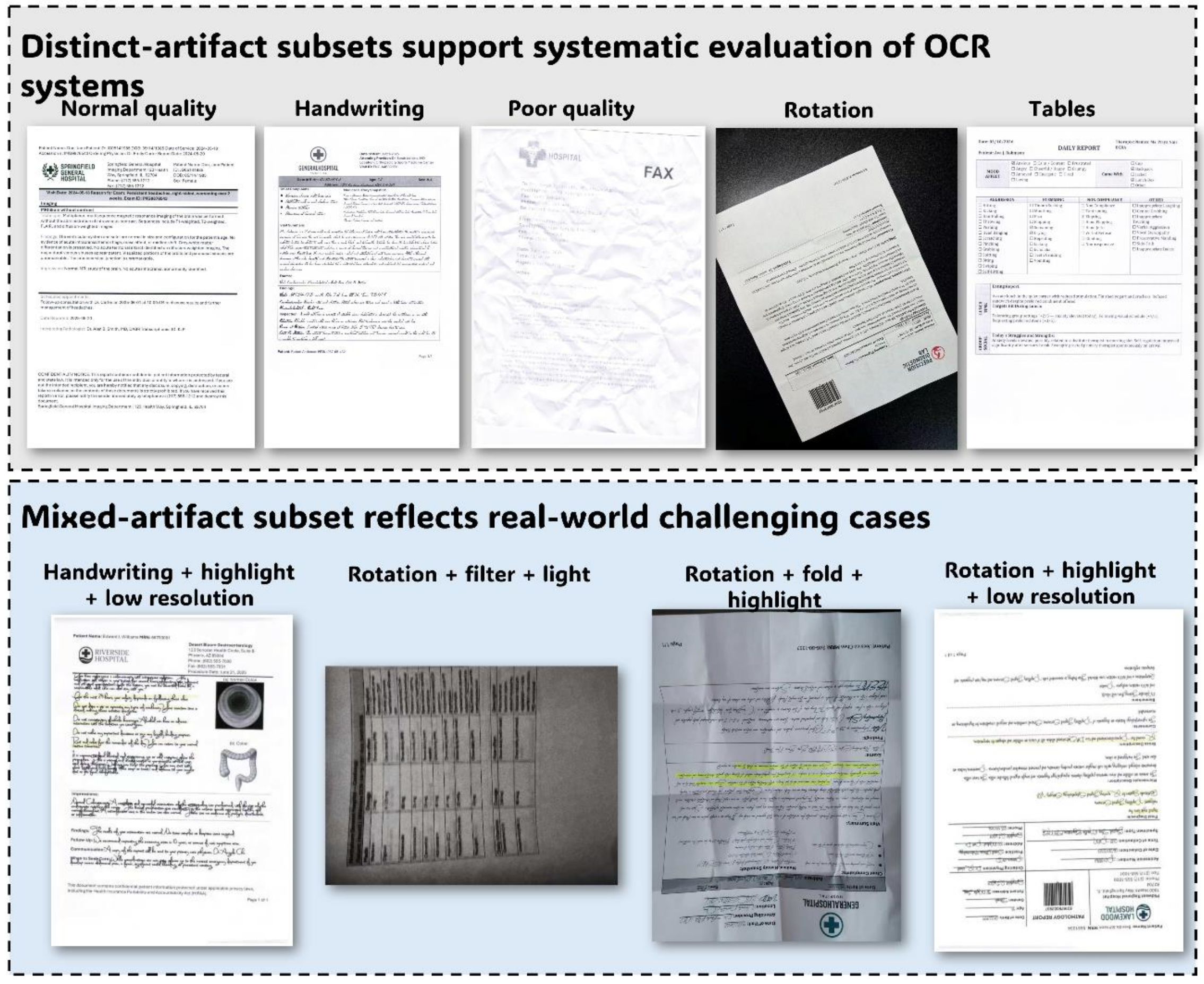

***Figure 1: Benchmark overview****. ClinOCR-Bench is comprised of 6 subsets: Normal, Handwriting, Poor Quality, Rotation, Tables, and Mixed, reflecting real-world OCR challenges while supporting a systematic OCR model evaluation.*

## Methods

The workflow of creating the synthetic scanned documents involves six steps: 1) **template creation**, which involves reviewing multiple scanned documents in the EHR and crafting blank layouts with text and figure regions (e.g., patient information region, hospital logo region, and lab findings region). 2) **Embedded images creation**, which involves prompting ChatGPT for hospital logo, barcode, and medical image generation. We post-processed those images (e.g., size adjustment and background removal) and placed them in the templates. 3) **Textual content generation**, which prompts LLMs (i.e., ChatGPT, Gemini 3 Pro, and Claude Sonnet 4.6) to generate medical text for each region and tabular content for tables. 4) **Layout and font adjustments**, which involve manual adoptions of the generated text into the document layouts (i.e., text regions and tables). Handwriting fonts were adopted for the "Handwriting" and "Mixed" subsets. Note that the "Tables" subset also involves handwriting fonts for provider signatures. 5) **Scanning artifact adoption**, in which each sample is printed out and adopted physical and programmatic degradation methods, including paper crumpling, multi-generation copies, and resolution reduction. 6) **Audit and revision**, in which co-authors review, comment, and revise the dataset through discussion (Figure 2). This section describes the methodology and provides references to our online materials.

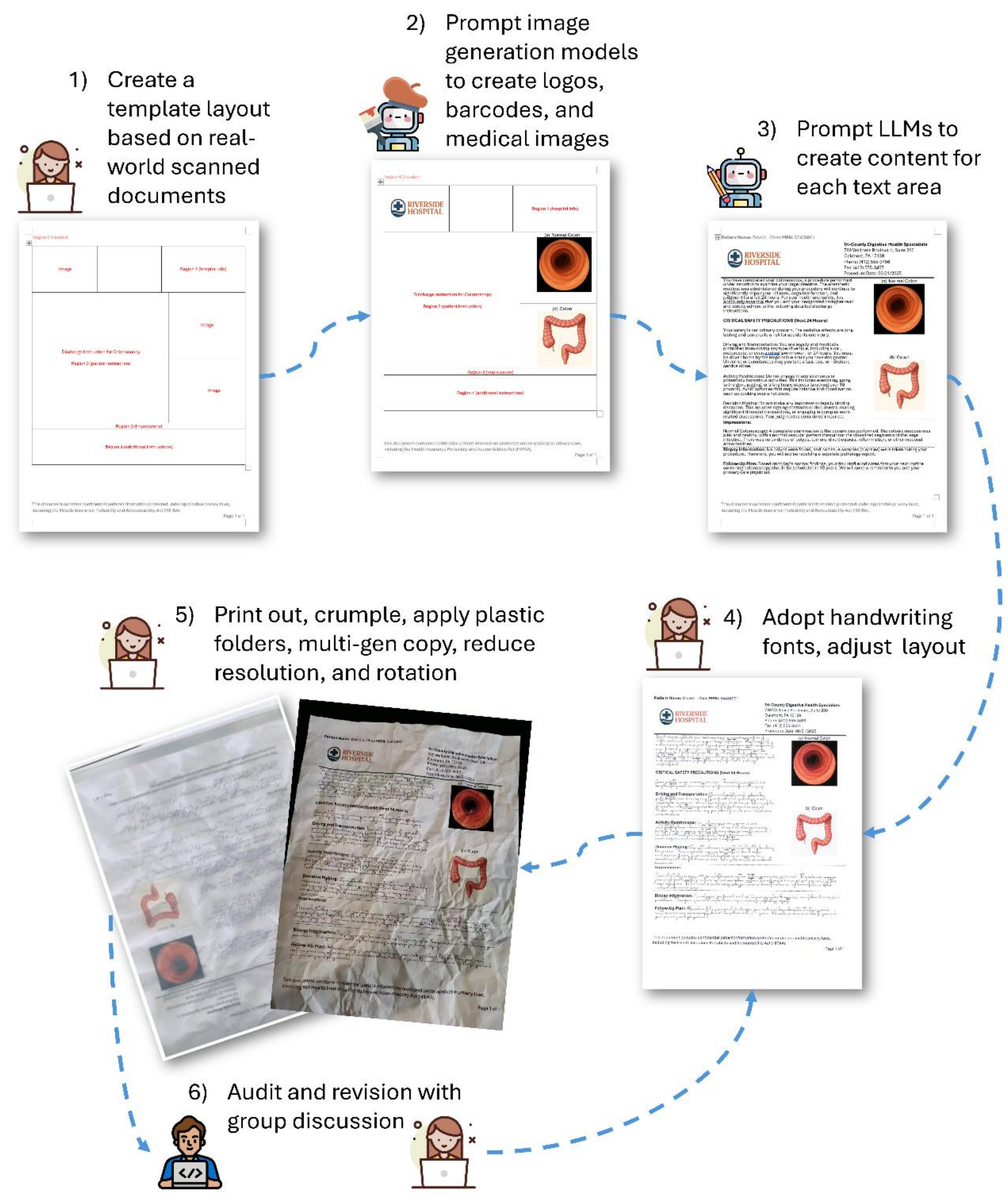

***Figure 2: Methodology overview**. 1) Manually create empty layouts inspired by medical documents in EHR. 2) Generate embedded images using image-generation AI tools. 3) Prompt generative AI to create textual content for each text region. 4) Adopt handwriting fonts to the documents and manually adjust layouts. 5) Print out the document and degrade the quality to mimic scanned documents in EHR. 6) Manual audit and revision to ensure data quality.*

**Templates**

We manually created document templates with Microsoft Word. Templates 1-8 cover non-tabular layouts for imaging report, laboratory report, pathology report, colonoscopy report, clinical visit note, patient summary, pathology report, and referral letter. In general, a template includes 4 to 12 text regions, page headers, footers, and a few placeholders for icons and images. Templates 9-16 cover tabular layouts for medication list, comprehensive metabolic panel, lipid panel, patient intake form, surgical pre-operative assessment, complete blood count, and behavioral session note. Page headers, footers, signature area, and placeholders for icons and images are included. All templates are available on our GitHub repository (https://github.com/ClinOCR-Bench/ClinOCR-Bench/tree/main/templates).

**Embedded images**

We prompted ChatGPT to generate hospital icons, colonoscopy images, and laboratory barcodes with the following prompts:

> *"Create a fake logo for a hospital that does not exist."*
>
> *"Create a fake colonoscopy image that looks like ones in lab reports"*
>
> *"Create a fake barcode with white background as those seen on lab orders."*

We then post-processed the images by adjusting sizes and removing backgrounds with Microsoft Paint 3D. All images, before and after adjustment, are available on our GitHub repository (https://github.com/ClinOCR-Bench/ClinOCR-Bench/tree/main/embedded_images).

**Textual and tabular content**

We prompt ChatGPT, Gemini 3 Pro, and Claude Opus 4.7 web application to generate the textual and tabular contents with the following prompt:

> *"Generate a new Word file (.docx) with synthesized data in the tables and the Regions (in red). Do not change the layout and font. The purpose is to create a real-looking medical report."*

We encountered diverse, unexpected outputs, including missing regions and unsuitable content. Multiple rounds of conversation were needed to achieve the final version. The generated content was then applied to the templates to create samples. All samples are available in Word (.docx) format on our GitHub repository (https://github.com/ClinOCR-Bench/ClinOCR-Bench/tree/main/samples).

**Handwriting fonts**

We applied 61 unique handwriting fonts to the samples to create the *handwriting* subset. The full list of fonts is available on our GitHub repository (https://github.com/ClinOCR-Bench/ClinOCR-Bench/blob/main/index.csv).

**Scanning artifacts**

We define rotation as skewness exceeding 30 ° from the regular document orientation. Note that minor skewness on pages or text areas (e.g., skewed handwritten text) is not considered rotation. **Rotation** was applied by photographing with a large angle greater than 30° or scanning with 90°, 180°, or 270° rotation. **Poor-quality** scanning involves a photograph taken with a phone camera at a vertical angle, with objects (e.g., fingers and table) in the background, or with poor lighting (e.g., sunlight from behind the paper or lamp from a corner of the page). It also involves scanning with folded or crumpled paper, marks on paper (e.g., highlighter), scanning through paper protector sheets, and multi-generation copy (i.e., copy on copies). Programmatic methods are adopted, including resolution reduction (e.g., reduce from 1700 to 600 pixels) and filters (e.g, *Instant*, *Moriyama*, and *BlackBerry 9000* in Meitu). The full list of artifacts with parameters is available on our GitHub repository (https://github.com/ClinOCR-Bench/ClinOCR-Bench/blob/main/index.csv).

**Audit and revision**

Four rounds of auditing were made before the data release. The main issues identified during the audit process were a) use of handwriting fonts on contents that should be printed (e.g., headers), b) textual content left out of the photograph, c) rotation angle less than 30°, and d) mistakes in document naming and matching. The complete audit log is available on our GitHub repository (https://github.com/ClinOCR-Bench/ClinOCR-Bench/blob/main/audit_log.txt).

## Data Records

ClinOCR-Bench includes 384 synthetic scans (56 exemplars for few-shot learning and 328 test images for evaluation) across 6 subsets, derived from 16 templates (i.e., document layouts with placeholders for text regions and tables), which covers 12 document types (i.e., imaging report, pathology report, laboratory report, colonoscopy report, clinical visit note, patient summary, diagnostic report, referral letter, medication list, patient intake form, surgical pre-operative assessment, and behavioral session note). Documents contain on average 200.0 words (SD = 72.4), with per-template means ranging from 111.2 (SD = 2.4) to 322.0 (SD = 66.7) (Table 1). Each template is used to create 8 samples (i.e., textual content and table values). The *Normal*, *Handwriting*, *Poor Quality*, and *Rotation* subsets are based on the same 64 samples from templates 1 to 8, while adopting different artifacts. The *Normal* subset includes scans without scanning artifacts; the *Handwriting* subset includes scans with the majority of content in handwriting fonts (e.g., Rembank, Sweetly Broken, and Quentin); the *Poor Quality* subset adopts physical degradation (e.g., folding and crumpling, photograph, and multi-generation copies) and programmatic degradation (e.g., reduced resolution and image filters) methods; the *Rotation* subset includes scans that are at least 30° rotated from the normal orientation. The *Tables* subset includes 64 samples, derived from templates 9 to 16, with the majority of their contents in complex tabular structures. The *Mixed* subset reuses 4 samples randomly chosen from each of the templates 1 to 16, while adopting multiple artifacts (e.g., handwriting, rotation, and reduced resolution on the same image) (Figure 3). The full description of each subset is available in our GitHub repository (https://github.com/ClinOCR-Bench/ClinOCR-Bench/blob/main/index.csv).

ClinOCR-Bench is delivered with two one-shot prompting configurations: 1) the **homogeneous setting**, where the one-shot exemplar and the test image share the same template. For example, the "template_1_sample_1_handwriting" document is used as the one-shot exemplar for the "template_1_sample_2_handwriting" document. This reflects a real-world scenario in which a patient form or laboratory report template is repeatedly used for multiple patients. 2) The **heterogeneous setting**, where the one-shot exemplar and the test image do not share the same template. This reflects scenarios in which scanned documents are received from heterogeneous sources and do not share a common layout. We adopted a cyclic design that assigns an exemplar from a different template to each test image. For example, when evaluating "template_1_sample_2", "template_2_sample_1" is used as the exemplar (Table 2). The full mapping is available on the GitHub page (https://github.com/ClinOCR-Bench/ClinOCR-Bench/blob/main/oneshot_lookup.csv).

*Table 1: Scanned document types and ground truth summary by template*

| Template | Document type | Word count, Mean (Std Dev) | Word count, Median [Q1, Q3] |
|---|---|---|---|
| Template 1 | Imaging/laboratory report | 296.0 (32.3) | 300 [288, 308] |
| Template 2 | Pathology report | 275.6 (38.8) | 283 [250, 288] |
| Template 3 | Colonoscopy report | 322.0 (66.7) | 333 [266, 375] |
| Template 4 | Clinical visit note | 310.1 (59.2) | 336 [256, 359] |
| Template 5 | Patient summary | 143.4 (11.6) | 146 [136, 152] |
| Template 6 | Pathology report | 228.4 (36.1) | 246 [190, 255] |
| Template 7 | Diagnostic report | 176.4 (17.2) | 169 [165, 180] |
| Template 8 | Referral letter | 213.4 (11.1) | 216 [202, 222] |
| Template 9 | Medication list | 151.1 (22.3) | 155 [141, 162] |
| Template 10 | Medication list | 117.1 (4.5) | 118 [116, 120] |
| Template 11 | Laboratory report (Comprehensive metabolic panel) | 168.0 (1.5) | 168 [167, 169] |
| Template 12 | Laboratory report (Lipid panel) | 127.2 (6.8) | 126 [123, 128] |
| Template 13 | Patient intake form | 164.2 (2.8) | 164 [162, 166] |
| Template 14 | Surgical pre-operative assessment | 213.1 (5.0) | 213 [210, 217] |
| Template 15 | Laboratory report (complete blood count) | 111.2 (2.4) | 112 [109, 113] |
| Template 16 | Behavioral session note | 182.4 (6.0) | 182 [177, 185] |
| Overall | | 200.0 (72.4) | 178 [150, 241] |

*Table 2: One-shot exemplar and test image mapping illustration*

| Sample | Role | Homogenous exemplar | Heterogeneous exemplar |
|---|---|---|---|
| Template 1 Sample 1 | Exemplar | N/A | N/A |
| Template 1 Sample 2 | Test | Template 1 Sample 1 | Template 2 Sample 1 |
| Template 1 Sample 3 | Test | Template 1 Sample 1 | Template 3 Sample 1 |
| Template 1 Sample 4 | Test | Template 1 Sample 1 | Template 4 Sample 1 |
| Template 1 Sample 5 | Test | Template 1 Sample 1 | Template 5 Sample 1 |
| Template 1 Sample 6 | Test | Template 1 Sample 1 | Template 6 Sample 1 |
| Template 1 Sample 7 | Test | Template 1 Sample 1 | Template 7 Sample 1 |

| Template 1 Sample 8 | Test | Template 1 Sample 1 | Template 8 Sample 1 |
|---|---|---|---|
| Template 2 Sample 1 | Exemplar | N/A | N/A |
| Template 2 Sample 2 | Test | Template 2 Sample 1 | Template 3 Sample 1 |
| Template 2 Sample 3 | Test | Template 2 Sample 1 | Template 4 Sample 1 |
| Template 2 Sample 4 | Test | Template 2 Sample 1 | Template 5 Sample 1 |
| Template 2 Sample 5 | Test | Template 2 Sample 1 | Template 6 Sample 1 |
| Template 2 Sample 6 | Test | Template 2 Sample 1 | Template 7 Sample 1 |
| Template 2 Sample 7 | Test | Template 2 Sample 1 | Template 8 Sample 1 |
| Template 2 Sample 8 | Test | Template 2 Sample 1 | Template 1 Sample 1 |
| … | | | |

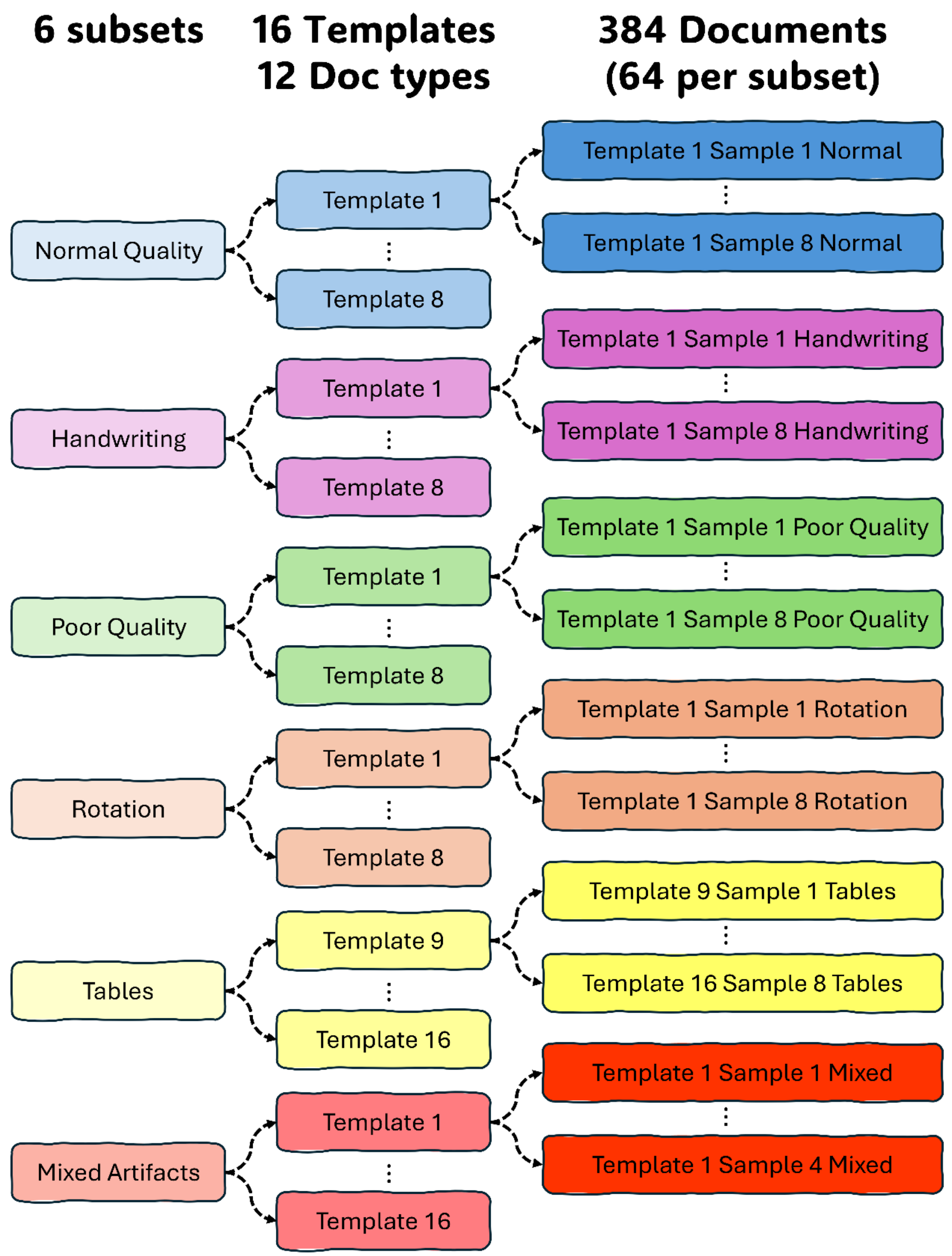

***Figure 3: File structure****. The ClinOCR-Bench dataset is derived from 16 templates. Each template is used to create 8 samples. The Normal, Handwriting, Poor Quality, and Rotation subsets share the same templates (templates 1 to 8) and samples, while the Tables subset is derived from*

*templates 9 to 16. The Mixed subset includes 64 samples from the 16 templates (4 samples per template).*

ClinOCR-Bench is publicly available, under MIT license, on both GitHub (https://github.com/ClinOCR-Bench/ClinOCR-Bench) and Hugging Face Hub (https://huggingface.co/datasets/ClinOCR-Bench/ClinOCR-Bench). The raw image files, ground truth, and one-shot exemplar mapping can be downloaded from the GitHub releases. To programmatically access the dataset, we recommend using the Hugging Face Hub Python package (Figure 4).

```python
from datasets import load_dataset

# pick a subset: normal | handwriting | poor | rotated | tables | mixed
ds = load_dataset("ClinOCR-Bench/ClinOCR-Bench", "handwriting")
test, exemplars = ds["test"], ds["train"]

# one-shot demos are referenced by id; resolve against the train split
by_id = {row["doc_id"]: row for row in exemplars}
item = test[0]
homo_demo = by_id[item["homo_id"]]      # same-template exemplar
hetero_demo = by_id[item["hetero_id"]]  # sibling-template exemplar

# item["image"], item["ground_truth"] are the query; *_demo carry the shots
```

***Figure 4: Python code to access ClinOCR-Bench from Hugging Face Hub.*** *The Hugging Face datasets Python package provides easy access to download and manage the dataset.*

## Technical Validation

### Methods to evaluate baseline OCR performance

To establish baseline OCR performance on ClinOCR-Bench, we conducted experiments with a traditional OCR engine, OCR-specialty VLMs, general-purpose open-weight VLMs, and proprietary multi-modal LLMs.

**OCR models**. We adopted the *Tesseract* OCR engine [17], a legacy computer vision-based toolkit, to represent non-VLM solutions. Two small VLMs (around 1B parameters), PaddleOCR-VL-1.5 [10] and LightOnOCR-2-1B [11], were adopted to represent OCR-specialty models. We adopted two general-purpose VLMs, *Qwen3.5-35B-A3B* [18] and *Gemma4-26B-A4B-it*. We also included proprietary multi-modal LLMs, *Gemini 3.5 flash,* and *GPT-5.4-mini*.

**Computing**. The open-weight models were deployed on a server with 4 NVIDIA RTX 3090 GPUs, using the *vLLM* inference engine version 0.20.1. The proprietary models were assessed through OpenRouter. All experiments were conducted with the *VLM4OCR* Python package version 0.5.0.

**Experimental design**. We experimented with both zero-shot and one-shot OCR. In the **zero-shot** setting, OCR models were given a 1) system prompt (except for Tesseract, PaddleOCR-VL-1.5, and LightOnOCR-2-1B), 2) a user prompt (except for Tesseract and LightOnOCR-2-1B), and 3) the target scanned document as an image (.jpg file). In the **one-shot** setting, an additional message with a one-shot example image and the ground truth ("exemplar") was provided before the target document image.

The system prompt is:

> *"You are a helpful assistant that can convert scanned documents into plain text. Your output is accurate and well-formatted. You will only output the plain text without any additional explanations or comments. The plain text should include all text, tables, and lists with appropriate layout (e.g., \n, space, and tab). You will ignore images, icons, or anything that can not be converted into text."*

The user prompt is:

> *"The input is a scanned medical document. Transcribe all of its content into plain text, preserving the reading order. Output only the transcribed text, with no commentary, explanations, or formatting markup."*

The full list of prompts is available on our GitHub page. Note that for *PaddleOCR-VL-1.5*, we used "OCR" as the user prompt instead, following the official guideline.

**Evaluation**. Word error rate (WER) is a standard metric for OCR performance [19]. It is defined as,

$$WER = \frac{i_w + s_w + d_w}{n_w},$$

Where $i_w$ is the number of words falsely inserted ("hallucinated"); $s_w$ is the number of words falsely substituted by the model; $d_w$ is the number of words falsely deleted ("omitted"); and $n_w$ is the number of words in the ground truth. WER ranges from $[0, \infty)$, with lower values indicating better OCR results. Note that WER can be greater than 1.0 when the OCR model hallucinates many words (i.e., $i_w + s_w + d_w > n_w$).

**Baseline OCR performance**

On the *Normal*, *Handwriting*, *Poor Quality*, and *Rotation* subsets, the *Qwen3.5-35B-A3B* with homogeneous one-shot prompting achieved the lowest median word error rate (WER) of 0.0000, 0.0315, 0.0070, and 0.0122, respectively. On the *Tables* subset, *Qwen3.5-35B-A3B*, *GPT-5.4-mini*, and *gemini-3.5-flash* with homogeneous one-shot achieved a median WER of 0.0000. On the *Mixed* subset, gemini-3.5-flash homogeneous one-shot dominates (median WER = 0.1636). Relative to the *Tesseract* baseline, VLMs showed dramatic improvements: *Normal* (0.0866 to 0.0000), *Handwriting* (0.8228 to 0.0315), *Poor Quality* (0.8184 to 0.0070), *Rotation* (1.0000 to 0.0122), *Tables* (0.4578 to 0.0000), and *Mixed* (1.0000 to 0.1636). Comparing the three prompting settings, median WER decreased monotonically from zero-shot to heterogeneous one-shot to homogeneous one-shot across most experiments. For *gemini-3.5-flash* on the *Mixed* subset, median WER decreased from 0.3495 (zero-shot) to 0.2949 (heterogeneous one-shot) to 0.1636 (homogeneous one-shot); for Qwen3.5-35B-A3B on *Tables*, the corresponding values were 0.1984, 0.0701, and 0.0000. Among the six subsets, the *Mixed* subset showed the highest WER under all configurations, with the majority of models retaining median WER above 0.4000 (Figure 4).

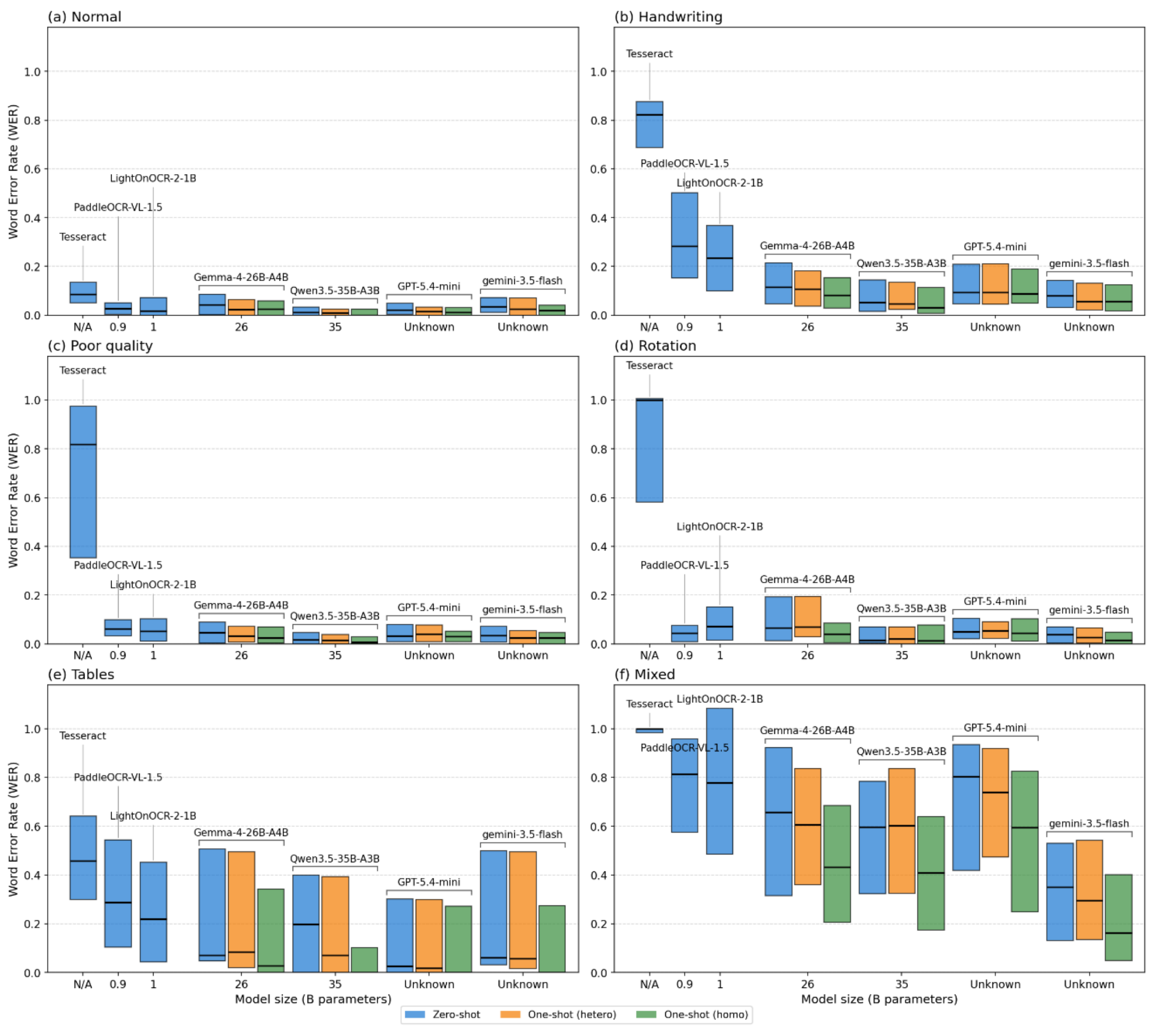

***Figure 5: Baseline OCR performance presented with median and first and third quartiles of word error rate (WER)****. Most VLMs show dramatic improvement from Tesseract. The Mixed subset showed the highest WER under all configurations, with the majority of models retaining a high median WER above 0.4.*

**Power calculation**

We performed an ad hoc power calculation to validate the sample size for each subset. Using the Wilcoxon signed rank test ($\alpha = 0.05$, two-sided), to achieve 80% power to distinguish the median WER of *Qwen3.5-35B-A3B* and *Tesseract*, it requires at least 26 scanned images (in the *Mixed* subset, N = 26, power = 0.9427; in the *Tables* subset, N=16, power=0.9999; for other subsets, N ≥ 6 results in power ≥ 0.8). Our sample size for each subset is adequate.

## Data Availability

ClinOCR-Bench is publicly available, under MIT license, on both GitHub (https://github.com/ClinOCR-Bench/ClinOCR-Bench) and Hugging Face Hub (https://huggingface.co/datasets/ClinOCR-Bench/ClinOCR-Bench).

## Code Availability

Our baseline experiment code, results, and documentation are available on GitHub (https://github.com/ClinOCR-Bench/ClinOCR-Bench-Baseline).

## Author Contributions

EH and KR conceived of the study and designed the baseline experiments. EH prototyped the methodology, designed the document templates, created the samples, and performed baseline experiments. JZ improved and conducted the data synthesis workflow (i.e., layout and fonts adoption, scanning artifacts adoption, and revision). All authors participated in the study design, data auditing, and manuscript writing. EH and JZ maintain the GitHub and Hugging Face Hub repositories.

## Funding

This work was supported by the National Library of Medicine grant numbers R01LM011934 and R01LM014508.